\documentclass[10pt, a4paper]{article}

\usepackage[final]{lrec2026} 

\usepackage{amssymb}
\usepackage{multicol}
\usepackage{url}
\usepackage{multirow}
\usepackage{booktabs}

\title{Do Lexical and Contextual Coreference Resolution Systems Degrade Differently under Mention Noise? An Empirical Study on Scientific Software Mentions}

\name{Atilla Kaan Alkan$^1$, Felix Grezes$^1$, Jennifer Lynn Bartlett$^1$,\\ \large \textbf{Anna Kelbert$^1$}, \textbf{Kelly Lockhart$^1$}, \large \textbf{Alberto Accomazzi}$^1$\\[0.05em]} 
\address{$^1$Harvard-Smithsonian Center for Astrophysics, Cambridge, MA, USA \\[0.05em]
         \{atilla.alkan, felix.grezes, jennifer.bartlett, anna.kelbert,\\ kelly.lockhart, alberto.accomazzi\}@cfa.harvard.edu\\}

\abstract{
We present our participation in the SOMD 2026 shared task on cross-document software mention coreference resolution, where our systems ranked second across all three subtasks. We compare two fine-tuning-free approaches: Fuzzy Matching (FM), a lexical string-similarity method, and Context Aware Representations (CAR), which combines mention-level and document-level embeddings. Both achieve competitive performance across all subtasks (CoNLL~F\textsubscript{1} of 0.94--0.96), with CAR consistently outperforming FM by 1 point on the official test set, consistent with the high surface regularity of software names, which reduces the need for complex semantic reasoning. A controlled noise-injection study reveals complementary failure modes: as boundary noise increases, CAR loses only 0.07 F\textsubscript{1} points from clean to fully corrupted input, compared to 0.20 for FM, whereas under mention substitution, FM degrades more gracefully (0.52 vs.\ 0.63). Our inference-time analysis shows that FM scales superlinearly with corpus size, whereas CAR scales approximately linearly, making CAR the more efficient choice at large scale. These findings suggest that system selection should be informed by both the noise profile of the upstream mention detector and the scale of the target corpus. We release our code to support future work on this underexplored task.
 \\ \newline  }

\begin{document}

\maketitleabstract

\section{Introduction}

Coreference resolution is the task of identifying all mentions in a text, including proper nouns, definite descriptions, pronouns, and nominal expressions that refer to the same real-world entity~\citep{jurafsky-etal-2026}. It is a core component of natural language understanding pipelines, underpinning downstream tasks such as information extraction~\citep{zelenko-etal-2004-coreference}, question answering~\citep{chai-etal-2022-evaluating}, and summarisation~\citep{huang-kurohashi-2021-extractive}. The automatic detection and disambiguation of software mentions in scientific literature has gained increasing attention as a means of improving research reproducibility and enabling large-scale meta-analyses of the use of methodologies across disciplines~\citep{schindler-etal-somesci,Schindler22:PeerJ,softcite}. 
In the scientific domain, and particularly for software mentions, coreference is predominantly nominal, with mentions taking the form of proper names, abbreviations, version-qualified strings, or URL references rather than pronouns or definite descriptions. This interest has been catalysed in part by shared evaluation campaigns: the SOMD 2025~\citep{upadhyaya-etal-2025-somd2025} shared task, focused on detecting software mentions and their semantic relations in scientific text, attracted a range of systems submissions~\citep{ojha-etal-2025-somd,rastogi-tiwari-2025-extracting,mandic-etal-2025-distribution,silva-etal-2025-inductive} and demonstrated both the feasibility and the remaining challenges of automated software mention extraction. Building on this foundation, SOMD 2026 extends the challenge to the cross-document setting, where the goal shifts from detecting individual mentions to resolving which mentions, across an entire collection of papers, refer to the same software entity. This is non-trivial: the same tool may appear under its full name, an acronym, a version-qualified string, or a URL, while conversely the same surface form may legitimately denote different tools in different disciplinary contexts.
While coreference resolution has been studied extensively in both within-document and cross-document settings for scientific text~\citep{chaimongkol-etal-2014-corpus,luan-etal-2018-multi,brack-etal,forer-etal}, these efforts have focused primarily on general scientific concepts, biomedical entities, and argumentative discourse structures. Software mentions, with their particular mix of proper names, versioned identifiers, and abbreviations, constitute a distinct entity type that has received no dedicated coreference treatment. SOMD 2026 thus serves as the first standardised benchmark for this task.

In this paper, we present our participation in the SOMD 2026 shared task and address the following research questions:

\begin{itemize}
    \item \textbf{RQ1} \textit{Can a simple, unsupervised lexical baseline compete with a contextual embedding approach on cross-document software
    coreference?} Given the high surface regularity of software names, we hypothesize that fuzzy string matching may constitute a strong baseline
    that is difficult to surpass without task-specific supervision.

    \item \textbf{RQ2} \textit{How robust are these approaches to different types and levels of annotation noise?} Real-world annotations are
    imperfect, and understanding system behaviour under controlled degradation provides insight into practical deployment limits.

    \item \textbf{RQ3} \textit{What is the precision--speed trade-off between the two approaches?} For large-scale literature mining pipelines, inference efficiency is a major concern alongside accuracy.
\end{itemize}

Our main contributions are: (i)~two competitive unsupervised baselines for the SOMD 2026 shared task; (ii)~as part of this work, a systematic noise injection study that, to our knowledge, is the first robustness analysis conducted in the context of software mention coreference resolution; and (iii)~a characterisation of the inference-time trade-off between lexical and neural approaches on this task. We release our code\footnote{\url{https://github.com/adsabs/SOMD-2026}} to support reproducibility and future work on this task.

The remainder of this paper proceeds as follows. Section~\ref{related_work} reviews related work and positions software mention coreference as a distinct problem. Section~\ref{task_corpus} describes the shared task, corpus, and training data analysis that informed our design choices. Section~\ref{sec:systems} presents our two systems. Section~\ref{sec:setup} describes the evaluation metrics and noise injection protocol. Section~\ref{sec:results} reports results and robustness findings. Section~\ref{sec:discussion} discusses broader implications and limitations. Section~\ref{sec:conclusion} concludes with a summary of key insights and directions for future research.

\section{Related Work}
\label{related_work}

\paragraph{Software Mention Detection and Disambiguation}
The extraction of software mentions from scientific text has received growing attention, with corpora such as \textsc{SoftCite}~\citep{softcite} and \textsc{SoMeSci}~\citep{schindler-etal-somesci} providing annotated mentions of software names, version numbers, URLs, and developer attributes. These efforts primarily focus on named entity recognition and attribute extraction rather than on coreference. Shared evaluation campaigns have further advanced the field: \citet{grezes-etal-2022-overview} organized the DEAL 2023 shared task on entity detection in astrophysics literature, which included software mentions as a target entity type. More recently, SOMD 2025~\citep{upadhyaya-etal-2025-somd2025} targeted the detection of software mentions and their relational attributes as named entities in scholarly text, attracting a range of submissions exploring detection and relation extraction \citep{ojha-etal-2025-somd,rastogi-tiwari-2025-extracting, mandic-etal-2025-distribution,silva-etal-2025-inductive}. Across these efforts, the focus has consistently remained on mention-level extraction; cross-document disambiguation and coreference resolution have received less dedicated treatment, a gap that SOMD 2026, to the best of our knowledge, is the first shared task to directly address.

\paragraph{Scientific Coreference Resolution}
Coreference resolution has been extensively studied in newswire and general-domain text \citep{lee-etal-2017-end,joshi-etal-2020-spanbert}, but scientific text presents distinct challenges: dense technical terminology, heavy use of abbreviations, and domain-specific entity types are poorly covered by general-purpose systems. Within-document coreference for scientific text has been widely addressed in the biomedical domain \citep{zweigenbaum:hal-00656514,lu-poesio-2021-coreference}, supported by annotated corpora such as \textsc{MedStract}~\citep{su-etal-geniamedco}, \textsc{Genia-MedCo}~\citep{su-etal-geniamedco}, and \textsc{DrugNerAR}~\citep{bedmar-etal-drugnerAR}. Coreference corpora have also been developed for broader scientific domains \citep{chaimongkol-etal-2014-corpus,brack-etal,luan-etal-2018-multi} and for astrophysics~\citep{alkan-etal-2024-enriching}. Cross-document scientific coreference has received comparatively less attention: \citet{Cattan2021SciCoHC} introduced \textsc{SciCo} for cross-document coreference of scientific concepts, while recent work has explored LLM-based relational reasoning \citep{forer-etal} and knowledge-graph-grounded entity linking \citep{dong-etal-kg} to improve cross-document resolution. Despite this growing body of work, software mentions, as a distinct entity type combining versioned identifiers and abbreviations, remain understudied.

\paragraph{Coreference Resolution Methods}
Early coreference research relied on unsupervised heuristics and rule-based approaches, drawing on linguistic constraints such as syntactic and semantic compatibility~\citep{HOBBS1978311,grosz-sidner-1986-attention, grosz-etal-1995-centering,haghighi-klein-2010-coreference}, with recent work demonstrating that simple unsupervised rules remain competitive in certain settings~\citep{stolfo-etal-2022-simple}. The field has since shifted toward supervised neural architectures for both within-document and cross-document resolution~\citep{clark-manning-2016-improving,Wiseman2015LearningAA, tourille-etal-2020-modele,gliosca-amsili-2019-resolution,barhom-etal-2019-revisiting, cattan-etal-2021}, and more recently started to explore zero-shot approaches leveraging pre-trained language models such as BERT~\citep{devlin-etal-2019-bert}, SciBERT~\citep{beltagy-etal-2019-scibert}, and LLMs with prompting strategies \citep{blevins-etal-2023-prompting,le-etal-2022-shot,le2023large}. However, these methods are designed for mention types that differ substantially from those in the SOMD 2026 shared task. Existing systems typically target pronominal anaphora and nominal expressions with high lexical diversity, whereas the annotation scheme adopted by the SOMD 2026 organisers focuses exclusively on explicit software name mentions (a mention type characterised by a high degree of surface form similarity between corefering expressions). This distinction matters: the linguistic variation that motivates complex neural architectures is largely absent here, making heavyweight models both unnecessary and costly. While lighter alternatives such as \textsc{FastCoref}~\citep{otmazgin-etal-2022-f} reduce computational overhead, they remain expensive for large-scale literature mining pipelines and are designed for general coreference rather than this specific mention type. The high surface regularity of software names and the need for scalable processing together motivate our choice of two lightweight, unsupervised approaches: fuzzy string matching, which directly exploits lexical similarity, and clustering over contextual embeddings, which captures semantic variation without fine-tuning.

\section{Task and Corpus}
\label{task_corpus}

\subsection{Task Definition}
\label{sec:task}

SOMD 2026 frames software mention disambiguation as a cross-document coreference resolution problem: given a set of software mention spans with their surrounding sentences and metadata, the goal is to partition all mentions into clusters such that each cluster corresponds to a single underlying software entity. The three subtasks differ in the quality of the input mentions and the scale of the corpus:

\begin{itemize}
    \item \textbf{Subtask 1} operates over gold-standard annotated mentions, providing an upper-bound evaluation of coreference resolution in isolation from mention detection errors;
    \item \textbf{Subtask 2} operates over automatically predicted mentions, reflecting real-world conditions where upstream mention detection is imperfect and introduces noise into the coreference input;
    \item \textbf{Subtask 3} operates over predicted mentions at a larger scale, explicitly targeting the computational efficiency challenge that arises as the volume of documents and the density of software name variants increase.
\end{itemize}

Our participation covers all three subtasks. Subtasks~2 and~3 operate on automatically predicted mentions, so the noise level in the input is inherent to the upstream mention detection pipeline and varies in ways that are difficult to quantify directly. To complement the official evaluation and gain a more controlled understanding of how input quality affects each system, we conduct a noise-injection study on the gold-standard training data, systematically varying the noise level across two perturbation types (\textbf{RQ2}). The scale dimension of Subtask~3 similarly motivates our inference-time analysis
(\textbf{RQ3}).

\subsection{Corpus Statistics}
\label{sec:data}

The shared task datasets comprise scholarly documents from scientific disciplines, annotated with software mention spans and their coreference chains. Two distinct training sets are provided: Subtask~1 uses gold-standard annotations, while Subtasks~2 and~3 share a training set of automatically predicted mentions. Table~\ref{tab:corpus_stats} reports corpus statistics for both.

\begin{table}[!h]
\centering
\scriptsize
\renewcommand{\arraystretch}{1.15}
\begin{tabular}{lrr}
\toprule
\textbf{Statistic} & \textbf{Subtask 1} & \textbf{Subtasks 2 \& 3} \\
                   & \textit{(gold)}    & \textit{(predicted)}     \\
\midrule
\multicolumn{3}{l}{\textit{Corpus}} \\
\quad Documents                              & 973    & 967   \\
\quad Sentences with mentions                & 2,153  & 2,140 \\
\quad Mention instances                      & 2,974  & 2,860 \\
\quad Unique surface forms                   & 837    & 791   \\
\midrule
\multicolumn{3}{l}{\textit{Coreference chains}} \\
\quad Total clusters                         & 733    & 699   \\
\quad Avg chain length                       & 4.06   & 4.09  \\
\quad Max chain length                       & 367    & 366   \\
\quad Singleton rate                         & 51.7\% & 52.5\% \\
\quad Cross-doc rate (all clusters)          & 20.6\% & 20.9\% \\
\quad Cross-doc rate (non-singletons)        & 42.7\% & 44.0\%  \\
\midrule
\multicolumn{3}{l}{\textit{Mention surface forms}} \\
\quad Avg surface forms / cluster            & 1.14   & 1.13  \\
\quad Avg intra-cluster lexical sim.         & 0.881  & 0.887 \\
\bottomrule
\end{tabular}
\caption{Corpus statistics for the SOMD 2026 training sets.
Subtask~1 uses gold-standard mentions; Subtasks~2 and~3 share
a common training set of automatically predicted mentions.
Cross-doc rate (non-singletons) excludes singleton clusters,
which require no linking decision.}
\label{tab:corpus_stats}
\end{table}

The two training sets are relatively similar across all statistics, suggesting that the automatic mention detector used for Subtasks~2 and~3 is of high quality: it recovers a comparable number of mentions (2,860 vs.\ 2,974), a similar coreference chain structure, and a nearly identical lexical similarity profile. The primary difference lies in the slightly lower mention count and number of unique surface forms, reflecting mentions that the detector failed to recover. This observation is consequential for our experimental design: since the real-world noise introduced by the upstream detector in Subtasks~2 and~3 is inherently mild and unquantified, we complement the official evaluation with a controlled noise injection study that systematically explores a wider range of noise levels, allowing us to characterise how each system degrades as input quality decreases (\textbf{RQ2}).

The coreference structure of both training sets reveals several properties that are consequential for system design. The clusters exhibit a highly skewed length distribution, with maximum chain lengths of 367 and 366 and average lengths of 4.06 and 4.09 for Subtasks~1 and 2\&3, respectively. This is consistent with a small set of high-frequency software names, such as \textsc{MATLAB}, dominating the corpus, whereas most tools appear infrequently. Notably, singleton rates of 51.7\% and 52.5\% indicate that over half of all mentions lack a coreferent counterpart, implying that any coreference system must be conservative in its linking decisions.

An analysis of the coreference chain structure reveals an important nuance regarding the cross-document nature of the task. The raw cross-document cluster rate is approximately 20\% in both training sets, suggesting that the resolution problem is predominantly within-document. However, this figure is strongly influenced by the high singleton rate of around 52\%: singletons trivially belong to a single document and require no linking decision. Among non-singleton chains, the cross-document rate rises to 42.7\% and 44.0\% for Subtasks~1 and 2\&3, respectively, confirming that the task poses a genuine cross-document disambiguation challenge for the majority of chains that actually require coreference linking.

Most consequentially for our system design choices, both training sets exhibit a high degree of lexical regularity within coreference chains. The average number of distinct surface forms per cluster is 1.14 and 1.13, respectively, indicating that coreferring mentions are almost always near-identical strings rather than paraphrases or pronominal references. This is confirmed by average intra-cluster lexical similarities of 0.881 and 0.887, substantially higher than what would be expected in general coreference corpora where chains mix proper names, nominal descriptions, and pronouns. This property directly motivates our choice of lightweight unsupervised approaches and supports the hypothesis underlying \textbf{RQ1} that lexical similarity alone may constitute a strong signal for this task.

\section{Systems}
\label{sec:systems}

\subsection{Fuzzy Matching}
\label{sec:fm}

The fuzzy matching system clusters software mentions based on the lexical surface similarity. For each pair of mention strings $m_i$ and $m_j$, we compute a similarity score $s(m_i, m_j) \in [0, 1]$ using the Ratcliff/Obershelp algorithm~\citep{RatcliffObershelp1988}, as implemented by \textsc{SequenceMatcher} in Python's \textsc{difflib} library. The Ratcliff/Obershelp algorithm computes similarity as twice the number of matching characters divided by the total number of characters in both strings, where matching characters are identified by recursively finding the longest common substring and applying the same procedure to the non-matching regions on either side. This makes it sensitive to the overall structure of the string rather than character-level edit distance, and well-suited to software names where shared substrings are a strong indicator of coreference (e.g., \textit{``GraphPad Prism''} and \textit{``GraphPad Prism 8''}). Two mentions are linked if $s(m_i, m_j) \geq \theta$, where the threshold $\theta$ is tuned on the training set. Clusters are then formed by applying transitive closure over all linked mention pairs, such that if $m_i$ is linked to $m_j$ and $m_j$ is linked to $m_k$, all three are assigned to the same cluster regardless of the direct similarity between $m_i$ and $m_k$. The fuzzy matching system operates solely on mention strings and is entirely agnostic to document context, mention type, and surrounding text. Its computational complexity is superlinear in the number of unique mention strings, which is manageable given the relatively small number of unique surface forms in the corpus (837 and 791 for Subtasks~1 and 2\&3, respectively; see Table~\ref{tab:corpus_stats}).

\subsection{Context Aware Representations}
\label{sec:car}

The context-aware representations system encodes each software mention using \textsc{all-MiniLM-L6-v2}~\citep{miniLM}, a lightweight sentence embedding model from the \textsc{Sentence Transformers} library. The model comprises 6 transformer layers and 22M parameters and has been trained to produce semantically meaningful sentence-level representations via knowledge distillation from larger models. Its compact size makes it well-suited to large-scale mention encoding without requiring GPU acceleration. Rather than encoding the mention in its sentential context alone, we separately encode two complementary signals and combine them into a single representation. First, the normalised mention string is encoded independently, producing a mention-level representation $\mathbf{e}_m \in
\mathbb{R}^d$ that captures the surface form of the software name. Second, a document-level representation $\mathbf{e}_d \in
\mathbb{R}^d$ is constructed by aggregating up to ten unique mention-bearing sentences from the same document into a single string, which is then encoded with \textsc{all-MiniLM-L6-v2}. This document context captures the broader thematic and disciplinary setting in which the mention appears, providing a complementary signal for cases where the same surface form refers to different software in different contexts. Both representations are independently normalised to unit length and combined as a weighted sum:

\begin{equation}
    \mathbf{e} = \alpha \cdot \mathbf{e}_m + (1 - \alpha) \cdot \mathbf{e}_d
\end{equation}

\noindent where $\alpha = 0.6$, giving slightly more weight to the mention-level signal. This design reflects the intuition confirmed by our corpus analysis (Section~\ref{sec:data}): software names are highly surface-regular, making the mention string the primary coreference signal, while document context provides a disambiguating signal for ambiguous cases. The combined representations are clustered using agglomerative clustering with cosine distance and average linkage. Since each mention is encoded independently by the sentence transformer without reference to other mentions, the encoding step scales linearly with corpus size; the subsequent agglomerative clustering step runs in $O(n^2 \log n)$ via \textsc{sklearn}'s precomputed distance matrix approach, avoiding the naive $O(n^3)$ complexity of a stored-matrix implementation.

\subsection{Threshold Tuning}
\label{sec:tuning}

The fuzzy matching system requires a single threshold hyperparameter $\theta$, defining the minimum Ratcliff/Obershelp similarity score above which two mentions are linked. We perform a grid search over $\theta \in [0, 1]$ on the training set, selecting the value that maximises CoNLL~F\textsubscript{1}. Since no development set is provided by the shared task, we use the full training set for this purpose. The selected threshold is $\theta = 0.83$ for Subtask~1 and $\theta = 0.84$ for Subtasks~2 and~3. The context-aware system uses a fixed distance threshold of $\delta = 0.4$ for the agglomerative clustering step, which was set empirically without formal tuning. For the noise-injection experiments, the threshold $\theta$ is re-tuned at each noise level using the same grid-search procedure, and the best achievable performance is reported for each noise condition. This provides an optimistic upper bound on the robustness of the fuzzy matching method, assuming that the system has access to a clean validation signal at each noise level. The implications of this choice are discussed further in Section~\ref{sec:limitations}.

\section{Experimental Setup}
\label{sec:setup}

\subsection{Evaluation Metrics}

All official test-set scores are computed by the shared task organisers using the standard coreference resolution metrics: MUC~\citep{Vilain1995AMC_MUC}, B$^3$ \citep{Bagga1998-B_cubed}, and CEAFe~\citep{luo-2005-coreference_ceaf}. The primary metric is CoNLL~F\textsubscript{1}~\citep{pradhan-etal-2014-scoring}, defined as the unweighted average of the three F\textsubscript{1} scores. In addition to coreference performance, we report the average inference time for each system in order to characterise the precision--speed trade-off between the two approaches (\textbf{RQ3}).

\subsection{Noise Injection Protocol}
As discussed in Section~\ref{sec:task}, the noise level introduced by the automatic mention detector in Subtasks~2 and~3 is inherent to the upstream pipeline and cannot be directly quantified. To complement the official evaluation, we conduct a controlled internal robustness analysis by injecting noise directly into the training set mentions and evaluating both systems on the resulting perturbed data. This diagnostic study is independent of the official test set and is not intended to be directly compared with the results in Table~\ref{tab:main}; its purpose is to assess how each system degrades as input quality decreases under controlled, quantifiable noise conditions (\textbf{RQ2}). The two noise types we consider are motivated by realistic failure modes of mention detection systems. \textbf{Boundary modification} simulates span boundary errors, which are among the most common annotation and detection mistakes in span-level tasks: a mention span is randomly extended or truncated, producing a slightly incorrect but plausible mention. \textbf{Mention substitution} simulates a context mismatch error, where a mention detection system correctly identifies a span as a software mention but associates it with the wrong software name: the mention string is replaced with a different software name sampled from the training set, preserving the syntactic structure of the sentence. Table~\ref{tab:noise_examples} illustrates each noise type on a concrete example. Each perturbation type is applied at different intensity levels: 0\%, 25\%, 50\%, 75\%, and 100\% of all mentions affected, and both systems are evaluated after each perturbation. 

\begin{table*}[!h]
\centering
\small
\begin{tabular}{@{}lp{12cm}@{}}
\toprule
\textbf{Noise Type} & \textbf{Example} \\
\midrule
\textbf{Original} &
We used \colorbox{blue!20}{MATLAB} for statistical analysis and
visualization. \\
\midrule
\textbf{Boundary} &
We used \colorbox{red!20}{MATLAB for} statistical analysis and
visualization. \\
\textbf{Modification} &
We used \colorbox{red!20}{the MATLAB} for statistical analysis and
visualization. \\
\midrule
\textbf{Mention} &
We used \colorbox{red!20}{Python} for statistical analysis and
visualization. \\
\textbf{Substitution} &
We used \colorbox{red!20}{NumPy} for statistical analysis and
visualization. \\
\bottomrule
\end{tabular}
\caption{Illustration of noise injection methods applied to a software mention. The original mention is shown in \colorbox{blue!20}{blue}, while noisy mentions are shown in \colorbox{red!20}{red}. Boundary modification extends or truncates mention spans; mention substitution replaces the mention string with another software name sampled from the training set. Each noise type simulates a different failure mode of upstream mention detection. All methods are tested at noise rates of 0\%, 10\%, 25\%, 50\%, 75\%, and 100\%.}
\label{tab:noise_examples}
\end{table*}

\section{Results}
\label{sec:results}

\subsection{Main Results on the Test Set}

Table~\ref{tab:main} reports the official test-set results for all participating SOMD 2026 systems alongside our two submissions. 

\begin{table*}[!h]
\centering
\small
\renewcommand{\arraystretch}{1.15}
\setlength{\tabcolsep}{4pt}
\begin{tabular}{lcccccccccccc}
\toprule
& \multicolumn{4}{c}{\textbf{Subtask~1}}
& \multicolumn{4}{c}{\textbf{Subtask~2}}
& \multicolumn{4}{c}{\textbf{Subtask~3}}
\\
\cmidrule(lr){2-5}\cmidrule(lr){6-9}\cmidrule(lr){10-13}
\textbf{System}
& CoNLL & B$^3$ & CEAF\textsubscript{e} & MUC & CoNLL & B$^3$ & CEAF\textsubscript{e} & MUC & CoNLL & B$^3$ & CEAF\textsubscript{e} & MUC \\
\midrule
System A$^\dagger$
  & \textbf{0.98} & \textbf{0.99} & \textbf{0.96} & \textbf{0.99} & \textbf{0.98} & \textbf{0.99} & \textbf{0.95} & \textbf{0.99} & \textbf{0.96} & \textbf{0.97} & \textbf{0.92} & \textbf{0.99} \\
System B$^\dagger$
  & 0.92 & 0.94 & 0.86 & 0.97 & 0.92 & 0.93 & 0.85 & 0.98 & $^\ddagger$ & $^\ddagger$ & $^\ddagger$ & $^\ddagger$ \\
\midrule
Fuzzy Matching
  & 0.95 & 0.95 & 0.91
  & \underline{0.98} & 0.95 & 0.94
  & 0.91 & \textbf{\underline{0.99}} & 0.94
  & 0.94 & 0.90 & \textbf{0.99} \\
Context Aware
  & \underline{0.96} & \underline{0.96} & \underline{0.93}
  & \underline{0.98} & \underline{0.96} & \underline{0.96}  & \underline{0.93} & \textbf{\underline{0.99}} & $^\ddagger$\
  & $^\ddagger$\ & $^\ddagger$\ & $^\ddagger$\ \\
\bottomrule
\end{tabular}%
\caption{Official test-set results for SOMD 2026. CoNLL is the average of MUC, B$^3$, and
CEAFe~F\textsubscript{1}. \textbf{Bold}\,=\,best overall. \underline{Underline}\,=\,best among our systems. $^\dagger$\,=\,other participants. $^\ddagger$\,=\,no submission on Codabench.}
\label{tab:main}
\end{table*}

The results reveal several findings. First, all systems, including our two unsupervised, fine-tuning-free baselines, achieve very high CoNLL~F\textsubscript{1} scores across all subtasks, ranging from 0.94 to 0.96. This confirms the hypothesis underlying \textbf{RQ1}: the high lexical regularity of software mention chains (avg.\ intra-cluster similarity of 0.881; see Table~\ref{tab:corpus_stats}) implies that even a simple surface-matching approach constitutes a strong baseline for this task. Second, our context-aware representations system consistently outperforms the fuzzy matching approach, achieving a CoNLL~F\textsubscript{1} of 0.96 on both Subtasks~1 and~2 compared to 0.95 for FM. While the absolute gap is modest (one point), it is consistent across all coreference metrics. The improvement is most visible on CEAF\textsubscript{e} (0.93 vs.\ 0.91), which penalises both over- and under-clustering and is therefore a more sensitive indicator of clustering quality than MUC, which is known to favour recall. This suggests that the document-level contextual representations in CAR provide a small but consistent benefit over purely lexical matching, likely by resolving ambiguous cases in which the same surface form can refer to different software across different disciplinary contexts. Third, both our systems outperform System~B$^\dagger$ on all subtasks by a margin of 3--4 CoNLL F\textsubscript{1} points, despite requiring no training or fine-tuning. This further underscores the strength of unsupervised lexical and lightweight embedding approaches for this task, and suggests that the added complexity of supervised systems may not yet be warranted given the current scale and lexical regularity of the corpus. Fourth, the performance of all systems is remarkably stable across Subtasks~1, 2, and~3. The transition from gold-standard mentions (Subtask~1) to automatically predicted mentions (Subtask~2) produces no measurable degradation for any system, which is consistent with our corpus analysis showing that the two training sets are nearly identical in their statistical properties (Table~\ref{tab:corpus_stats}). This suggests that the upstream mention detector used by the shared task organisers is of high quality.

\subsection{Robustness to Mention Noise}
\label{sec:noise_results}

We evaluate robustness by injecting controlled noise into the training sets of Subtasks~1 and~2. Table~\ref{tab:noise_results} reports degradation curves for both noise types. As noted in Section~\ref{sec:tuning}, FM threshold $\theta$ is re-tuned at each noise level, so FM scores represent an optimistic upper bound.

\begin{table*}[t]
\centering
\renewcommand{\arraystretch}{1.15}
\setlength{\tabcolsep}{4.5pt}
\begin{tabular}{lrcccccc}
\toprule
& & \multicolumn{5}{c}{\textbf{Injected Noise Rate}} & \\
\cmidrule(lr){3-7}
\textbf{Noise Type} & \textbf{System}
& \textbf{$\emptyset$} & \textbf{25\%}
& \textbf{50\%} & \textbf{75\%} & \textbf{100\%}
& \textbf{$\Delta$} \\
\midrule
\multicolumn{8}{l}{\textit{Subtask 1 (gold mentions)}} \\
\midrule
\multirow{2}{*}{Boundary Modification}
  & Fuzzy Matching  & 0.70 & 0.65 & 0.60 & 0.54 & 0.50 & 0.20 \\
  & Context Aware   & \underline{0.82} & \underline{0.80} & \underline{0.79} & \underline{0.77} & \underline{0.75} & \textbf{0.07} \\
\midrule
\multirow{2}{*}{Mention Substitution}
  & Fuzzy Matching  & 0.70 & 0.49 & 0.34 & 0.23 & 0.18 & \textbf{0.52} \\
  & Context Aware   & \underline{0.82} & \underline{0.57} & \underline{0.39} & \underline{0.25} & \underline{0.19} & 0.63 \\
\midrule
\multicolumn{8}{l}{\textit{Subtask 2 (predicted mentions)}} \\
\midrule
\multirow{2}{*}{Boundary Modification}
  & Fuzzy Matching  & 0.70 & 0.66 & 0.62 & 0.57 & 0.51 & 0.19 \\
  & Context Aware   & \underline{0.82} & \underline{0.80} & \underline{0.80} & \underline{0.78} & \underline{0.76} & \textbf{0.06} \\
\midrule
\multirow{2}{*}{Mention Substitution}
  & Fuzzy Matching  & 0.70 & 0.50 & 0.34 & 0.23 & 0.18 & \textbf{0.52} \\
  & Context Aware   & \underline{0.82} & \underline{0.59} & \underline{0.39} & \underline{0.26} & \underline{0.20} & 0.62 \\
\bottomrule
\end{tabular}
\caption{CoNLL~F\textsubscript{1} under all noise types at increasing noise rates on the training set. $\Delta$ = F\textsubscript{1} at 0\% $-$ F\textsubscript{1} at 100\% (lower = more robust). Baseline scores differ from official test-set results (Table~\ref{tab:main}) as noise experiments are conducted on the training set. \underline{Underline} = best F\textsubscript{1}; \textbf{bold} $\Delta$ = most robust system per noise type.}
\label{tab:noise_results}
\end{table*}

\paragraph{Boundary Modification}
FM loses 0.20 CoNLL~F\textsubscript{1} points over the full noise range (0.70 $\rightarrow$ 0.50) while CAR loses only 0.07 (0.82 $\rightarrow$ 0.75), a three-fold robustness advantage. FM degrades approximately linearly, while CAR's curve is nearly flat. This asymmetry reflects how each system processes mention strings: FM relies on exact character-level matching, so adding or removing one token directly degrades similarity scores. CAR encodes mentions as dense vectors, so minor boundary changes leave the semantic content largely intact (e.g., \textit{"MATLAB for"} encodes similarly to \textit{"MATLAB"}). This robustness is a property of the embedding representation itself, not evidence that document context provides a compensatory signal, as the mention substitution results confirm. Practically, CAR at 100\% boundary noise (0.75) still outperforms FM at 0\% noise~(0.70).

\paragraph{Mention Substitution}
The robustness ranking reverses: FM ($\Delta = 0.52$) is now more robust than CAR ($\Delta = 0.63$), despite starting from a lower clean baseline. Both systems converge to near-identical performance at 100\% noise (FM: 0.18, CAR: 0.19). FM degrades approximately linearly, whereas CAR collapses more sharply at low noise levels: at 25\% noise, CAR drops by 0.25 points, whereas FM drops by 0.21. This early collapse occurs because CAR's document context is constructed from mention-bearing sentences (Section~\ref{sec:car}): substituting mention strings corrupts both the mention-level and document-level embeddings simultaneously, so neither component can compensate for the other. This resolves the apparent tension with the boundary modification results: CAR's robustness advantage there reflects the tolerance of dense embeddings to minor perturbations, not an independent contextual signal. When mention content is substantively replaced, this advantage disappears.

\paragraph{Consistency and Summary}
Degradation patterns are consistent across Subtasks~1 and~2 (with differences of at most 0.01 across all conditions), confirming that the findings generalise to the predicted-mention setting. Overall, the two systems exhibit complementary robustness profiles: CAR is more robust to boundary noise while FM degrades more gracefully under mention substitution, and neither system is resilient to severe mention content corruption (\textbf{RQ2}).

\subsection{Inference Time and Precision–Speed Trade-off}
\label{sec:time_results}

We report in Table~\ref{tab:time} the inference time and precision--speed trade-off for both systems on the official test sets of Subtasks~1 and~2, measured on a CPU-only machine (Intel x86\_64, 14 physical cores, 33.3 GB RAM) over 5 runs.

\begin{table}[!h]
\centering
\scriptsize
\renewcommand{\arraystretch}{1.15}
\begin{tabular}{lrrrr}
\toprule
\textbf{Subtask} & \textbf{System} & \textbf{Time (s)} & \textbf{CoNLL F\textsubscript{1}} & \textbf{F1/sec} \\
\midrule
\multirow{2}{*}{Subtask 1}
  & FM     & 0.60 $\pm$ 0.00 & 0.95 & 1.584 \\
  & CAR & 4.45 $\pm$ 0.69 & 0.96 & 0.215 \\
\midrule
\multirow{2}{*}{Subtask 2}
  & FM    & 47.06 $\pm$ 0.41 & 0.95 & 0.020 \\
  & CAR & 45.39 $\pm$ 5.14 & 0.96 & 0.021 \\
\bottomrule
\end{tabular}
\caption{Inference time and precision--speed trade-off on official test sets of Subtasks 1 and 2. Efficiency\,=\,CoNLL~F\textsubscript{1}\,/\, inference time (s). Measured on an Intel x86\_64 CPU (14 cores, 33.3 GB RAM) over 5 runs.}
\label{tab:time}
\end{table}

At small scale (Subtask~1, $n$\,=\,743 mentions), FM is 7.4$\times$ faster than CAR (0.60s vs.\ 4.45s) at near-identical accuracy (0.95 vs.\ 0.96), making it the more efficient choice for small corpora. At large scale (Subtask~2, $n$\,$\approx$\,21,500 mentions), the picture reverses: the input size increases 29$\times$ relative to Subtask~1, but FM's inference time increases 78$\times$ (0.60s $\rightarrow$ 47.06s) while CAR's increases only 10$\times$ (4.45s $\rightarrow$ 45.39s), and both systems converge to nearly identical inference times ($\approx$46s) with CAR still holding its marginal accuracy advantage. This result is counter-intuitive: despite being the simpler and lighter system, FM scales less favourably than CAR at a large scale. The reason is that FM's simplicity relies on pairwise string comparison, whose cost grows faster than linearly with the number of mentions, whereas CAR's neural encoding (though more expensive per mention) processes each mention independently and therefore scales approximately linearly. FM's inference time is perfectly stable across runs ($\pm$0.00s), reflecting its deterministic procedure, whereas CAR exhibits higher variance ($\pm$0.69s on Subtask~1, $\pm$5.14s on Subtask~2), attributable to variability in the neural encoding step on the CPU; GPU acceleration would reduce both.

\section{Discussion}
\label{sec:discussion}

\paragraph{Accuracy vs.\ scale (RQ3)}
At small scale (Subtask~1, $n$\,=\,743), FM is 7.4$\times$ faster than CAR at near-identical accuracy, making it the more practical choice. At large scale (Subtask~2, $n$\,$\approx$\,21,500), this advantage disappears: FM's inference time increases 78$\times$ while CAR's increases only 10$\times$, with both systems converging to similar inference times ($\approx$46s). This is consistent with FM's superlinear scaling in pairwise similarity computation, compared with CAR's approximately linear per-mention encoding. The practical recommendation is scale-dependent: FM for small corpora, CAR for large-scale pipelines.

\paragraph{Robustness, lexical regularity, and system selection (RQ1 \& RQ2)}
A unifying finding across RQ1 and RQ2 is that the high lexical regularity of software mention chains, confirmed by an average intra-cluster similarity of 0.881 (Table~\ref{tab:corpus_stats}), is the dominant factor shaping system behaviour. It explains why FM is competitive with CAR on clean data, why CAR's robustness advantage under boundary noise reflects embedding tolerance rather than genuine contextual signal, and why both systems collapse equally under mention substitution. Above a certain noise threshold, improving the upstream detector is more impactful than the choice of coreference method. Future work would benefit from document-level representations genuinely decoupled from mention string content. For instance, encoding non-mention sentences or document signals such as titles and abstracts.

\paragraph{Limitations}
\label{sec:limitations}
Three limitations should be noted. First, FM's threshold $\theta$ was re-tuned at each noise level, so robustness results represent an optimistic upper bound. Second, CAR's document context is constructed from mention-bearing sentences, making it structurally dependent on mention string content. Third, inference times were measured on CPU only; GPU acceleration would reduce CAR's inference time and potentially widen its efficiency advantage at scale. Table~\ref{tab:guide} summarises these recommendations. The key insight is that neither system is universally superior: the right choice depends jointly on corpus scale and the error profile of the upstream mention detector.

\begin{table}[!h]
\centering
\small

\renewcommand{\arraystretch}{1.15}
\begin{tabular}{lc}
\toprule
\textbf{Deployment condition} & \textbf{Recommended} \\
\midrule
Small corpus, high detector quality   & FM  \\
Large corpus ($n \gg$ 1k mentions) & CAR \\
Boundary errors dominate              & CAR \\
Mention identity errors dominate      & Improve detector \\
\bottomrule
\end{tabular}
\caption{Practical system selection guide based on
deployment context.}
\label{tab:guide}
\end{table}

\section{Conclusion}
\label{sec:conclusion}

We presented two unsupervised, fine-tuning-free systems for cross-document software mention coreference resolution at SOMD 2026, ranking second across all three subtasks. Our results answer three research questions. \textbf{RQ1}: The high lexical regularity of software mention chains makes unsupervised lexical methods strong baselines, competitive with contextual approaches without any task-specific supervision. \textbf{RQ2}: The two systems exhibit complementary robustness profiles: CAR is more robust to boundary noise, while FM degrades more gracefully under mention substitution, and neither is resilient to severe mention content corruption, pointing to upstream mention detection as the primary bottleneck for future progress. \textbf{RQ3}: FM is more efficient at a small scale while CAR scales more favourably to large corpora, making scale a decisive factor in system selection. Taken together, these findings provide concrete guidance for practitioners (Table~\ref{tab:guide}): system selection should be informed by the upstream detector's noise profile and the scale of the target corpus, not by accuracy alone. We release our code\footnote{\url{https://github.com/adsabs/SOMD-2026}} and hope that the noise-injection protocol introduced here serves as a reusable, robustness-evaluation framework for this underexplored task.

\section{Bibliographical References}\label{sec:reference}

\bibliographystyle{lrec2026-natbib}
\bibliography{lrec2026-example}





\end{document}